\documentclass[12pt]{article}

\usepackage{authblk}
\newcommand*\samethanks[1][\value{footnote}]{\footnotemark[#1]}

\usepackage{caption}
\usepackage{booktabs}
\usepackage{array}
\usepackage{enumitem}
\usepackage{amsmath}
\usepackage{amssymb}
\usepackage{graphicx}
\usepackage{placeins}
\usepackage{geometry}
\usepackage{url}
\usepackage{xpatch}
\usepackage{color}
\usepackage{setspace}
\usepackage{longtable}

\usepackage[
  backend=biber,
  style=vancouver,
  citestyle=numeric-comp,
  sorting=none,
  sortcites=true,
  giveninits=true,
  maxnames=3,
  minnames=3,
  doi=true,
  url=true,
  isbn=false,
  date=year
]{biblatex}

\DeclareFieldFormat{citerangeseparator}{-}

\xpatchbibmacro{journal+issuetitle}
  {\setunit{\addsemicolon\space}}
  {\setunit{\addsemicolon}}
  {}{}

\DeclareFieldFormat[article]{journaltitle}{\mkbibemph{#1}}

\AtEveryBibitem{%
  \iffieldundef{pages}{}{}%
}

\title{\vspace{-2em} RELEAP: Reinforcement-Enhanced Label-Efficient Active Phenotyping for Electronic Health Records}
\date{} 
\author[1]{Yang Yang, BS}
\author[2,3]{Kathryn I. Pollak, PhD}
\author[1,4,5,6]{Bibhas Chakraborty, PhD}
\author[7,8]{Molei Liu, PhD\thanks{Co-last authors contributed equally to this work.}}
\author[6]{Doudou Zhou, PhD\samethanks[1]}
\author[1]{Chuan Hong, PhD\samethanks[1]}

\affil[1]{Department of Biostatistics and Bioinformatics, Duke University, Durham, NC, USA}
\affil[2]{Cancer Prevention and Control Research Program, Duke Cancer Institute, Durham, NC, USA}
\affil[3]{Department of Population Health Sciences, Duke University School of Medicine, Durham, NC, USA}
\affil[4]{Centre for Quantitative Medicine, Duke-NUS Medical School, Singapore}
\affil[5]{Programme in Health Services and Systems Research, Duke-NUS Medical School, Singapore}
\affil[6]{Department of Statistics and Data Science, National University of Singapore, Singapore}
\affil[7]{Department of Biostatistics, Peking University Health Science Center, Beijing, China}
\affil[8]{Beijing International Center for Mathematical Research, Peking University, Beijing, China}

\begin{document}
\maketitle

\textbf{Corresponding Author:} 

Chuan Hong

\newpage
\section*{ABSTRACT}
\textbf{Objective:}  
Electronic health record (EHR) phenotyping often relies on noisy proxy labels, which undermine the reliability of downstream risk prediction. Active learning can reduce annotation costs, but most rely on fixed heuristics and do not ensure that phenotype refinement improves prediction performance. Our goal was to develop a framework that directly uses downstream prediction performance as feedback to guide phenotype correction and sample selection under constrained labeling budgets.  

\textbf{Materials and Methods:}  
We propose Reinforcement-Enhanced Label-Efficient Active Phenotyping (RELEAP), a reinforcement learning–based active learning framework. RELEAP adaptively integrates multiple querying strategies and, unlike prior methods, updates its policy based on feedback from downstream models. We evaluated RELEAP on a de-identified Duke University Health System (DUHS) cohort (2014--2024) for incident lung cancer risk prediction, using logistic regression and penalized Cox survival models. Performance was benchmarked against noisy-label baselines and single-strategy active learning.  

\textbf{Results:}  
RELEAP consistently outperformed all baselines. Logistic AUC increased from 0.774 to 0.805 and survival C-index from 0.718 to 0.752. Using downstream performance as feedback, RELEAP produced smoother and more stable gains than heuristic methods under the same labeling budget.  

\textbf{Discussion:}  
By linking phenotype refinement to prediction outcomes, RELEAP learns which samples most improve downstream discrimination and calibration, offering a more principled alternative to fixed active learning rules.  

\textbf{Conclusion:}  
RELEAP optimizes phenotype correction through downstream feedback, offering a scalable, label-efficient paradigm that reduces manual chart review and enhances the reliability of EHR-based risk prediction.

\textbf{Keywords:} Reinforcement Learning; Active Learning; Phenotyping; Risk Prediction; Lung Cancer.

\newpage
\section{INTRODUCTION}
Phenotyping refers to the process of deriving clinically meaningful traits, such as diseases, behaviors, or risk factors, from raw health data using structured codes, laboratory results, medications, and unstructured clinical text. Intermediate phenotypes play a central role in biomedical prediction models, serving as covariates or mediators that connect underlying patient characteristics to clinical outcomes. Their accuracy directly shapes the performance, interpretability, and fairness of downstream risk prediction. Within electronic health records (EHRs), however, such phenotypes are rarely observed directly. Instead, they must be inferred from raw data sources, including diagnoses, procedures, laboratory results, medications, and clinical text, using computable phenotypes, algorithmically defined constructs based on EHR data (e.g., diagnosis codes, medications, laboratory values, or unstructured clinical text) that approximate clinical conditions or behaviors.\cite{denny2010phewas,zhang2019high,shivade_review_2014}

Developing high-quality computable phenotypes at scale remains a major bottleneck. While unsupervised approaches like clustering and rule-based heuristics avoid the need for labels, they often lack specificity and clinical validity.\cite{liao2019high,halpern_anchor_2016} Supervised learning offers higher accuracy but depends on manually curated labels derived from chart review, which is time-intensive, costly, and unscalable. These challenges have sparked growing interest in label-efficient phenotyping strategies, which aim to reduce annotation burden without compromising phenotype quality. Approaches such as weak supervision,\cite{nogues2022weakly} surrogate labeling,\cite{hong2025label} and active learning exemplify this shift toward minimizing reliance on gold-standard labels.\cite{chen2013applying}

The accuracy of label-efficient phenotyping can be significantly enhanced by leveraging multiple data modalities. Structured EHR elements, such as International Classification of Diseases (ICD) codes,\cite{hong2023international} are often used as proxies for behavioral phenotypes like smoking, but their sensitivity is low and subject to substantial label noise.\cite{wiley_icd9_2013} For instance, studies have shown that relying solely on structured codes captures only a small subset of true smokers and systematically underrepresents smoking prevalence.\cite{ruckdeschel2023unstructured} Incorporating unstructured clinical text through natural language processing (NLP) and integrating self-reported data from social history or patient intake forms provides a more complete and accurate phenotype. Such multi-modal integration improves coverage and reduces misclassification, forming a more robust foundation for label-efficient learning.\cite{rajendran_nlp_smoking_2020}

A further complication is that while small validation cohorts from chart review can be used to benchmark
algorithms, such labels are insufficient for monitoring performance across diverse populations or for guiding iterative refinement. This limitation motivates the use of downstream prediction performance to guide active learning rather than relying exclusively on scarce gold-standard labels. A related line of work has sought to directly connect phenotyping with risk prediction. For example, Hong et al.\ proposed a semi-supervised framework that jointly
validates multiple surrogate phenotypes while estimating associations with genetic risk factors.\cite{hong2019semi} By combining the phenotyping step with the downstream risk genetic association study, their method improved both phenotype accuracy and the reliability of subsequent genetic association analyses. This illustrates the value of integrating phenotype refinement with predictive modeling, though existing approaches have largely focused on semi-supervised inference rather than adaptive sample selection.

To facilitate consistent terminology, we define the \textbf{automatic reference phenotype} as a high-fidelity label that integrates structured self-reported smoking information with large language model-based extraction from clinical notes. This multimodal phenotype serves as the most accurate available reference for downstream evaluation and is denoted as $S_{\text{true}}$ throughout this paper.

Building on this definition, we present \textbf{Reinforcement-Enhanced Label-Efficient Active Phenotyping (RELEAP)}, a reinforcement learning-driven active learning agent designed to improve phenotype correction under constrained labeling budgets.\cite{settles2012active,fang_learning_2017} Unlike prior label-efficient approaches, RELEAP directly uses downstream risk prediction performance as a feedback signal for phenotype refinement. By leveraging automatically constructed reference labels instead of manual chart review, RELEAP is both scalable and cost-efficient. Building on prior work such as Hong et al.,\cite{hong2019semi} we benchmark RELEAP against noisy-label and single-strategy baselines and quantify label efficiency by the labeling budget required to approach near-oracle performance.

\section{MATERIALS AND METHODS}
\subsection{Motivation Example: Lung Cancer Prediction with Imprecise Smoking Phenotypes}
Lung cancer remains a leading cause of cancer-related mortality, and risk prediction models are increasingly being developed to support earlier detection and preventive care. Recent machine learning models trained on large-scale EHR data have demonstrated strong performance in identifying high-risk patients, achieving area under the receiver operating characteristic curve (AUC) values above 0.75 across diverse real-world settings.\cite{chandran2023machine} In these models, smoking behavior consistently emerges as one of the most important predictors, alongside age, race, and chronic respiratory conditions such as chronic obstructive pulmonary disease (COPD).\cite{usptf_lungcancer_2021}

However, accurately capturing smoking status in EHR data is challenging. In routine care, smoking behavior is often recorded inconsistently, if at all. Proxy variables (e.g., diagnosis codes related to tobacco use) are commonly used for computable phenotyping, but their reliability is questionable. For example, in our analysis of EHR data from the Duke University Health System (DUHS), we found that the ICD-based proxy smoking phenotype ($S^{\ast}$) is markedly incomplete when compared to self-reported smoking status. While structured self-report fields in the vital signs table indicate that approximately 30\% of patients have a documented history of smoking, only about 10\% of patients carry any smoking-related ICD-9/10 code (e.g., \texttt{305.1}, \texttt{F17.*}, \texttt{Z87.891}). This large discrepancy highlights the limitations of relying on structured diagnosis codes as a proxy for smoking behavior. The sparse and imbalanced coverage of ICD codes fails to capture the true prevalence of smoking in the population and introduces substantial misclassification bias when used in downstream risk prediction models.

To address this gap, we constructed an automatic reference phenotype by combining structured self-reported smoking information with unstructured smoking mentions extracted from clinical notes using large language models prior to the prediction index date. This multimodal definition provides a more complete and accurate depiction of patient smoking history and serves as a reference label for evaluation. The discrepancy between $S^{\ast}$ and $S_{\text{true}}$ highlights a budget-constrained decision problem: given limited access to high-fidelity reference labels, determining which patients should be upgraded from noisy proxies in order to maximize downstream prediction performance. This decision problem motivates the design of RELEAP, which adaptively allocates scarce label queries rather than relying on fixed heuristics.

\subsection{Proposed RELEAP agent}
\begin{figure}[h]
    \centering
    \includegraphics[scale=0.75]{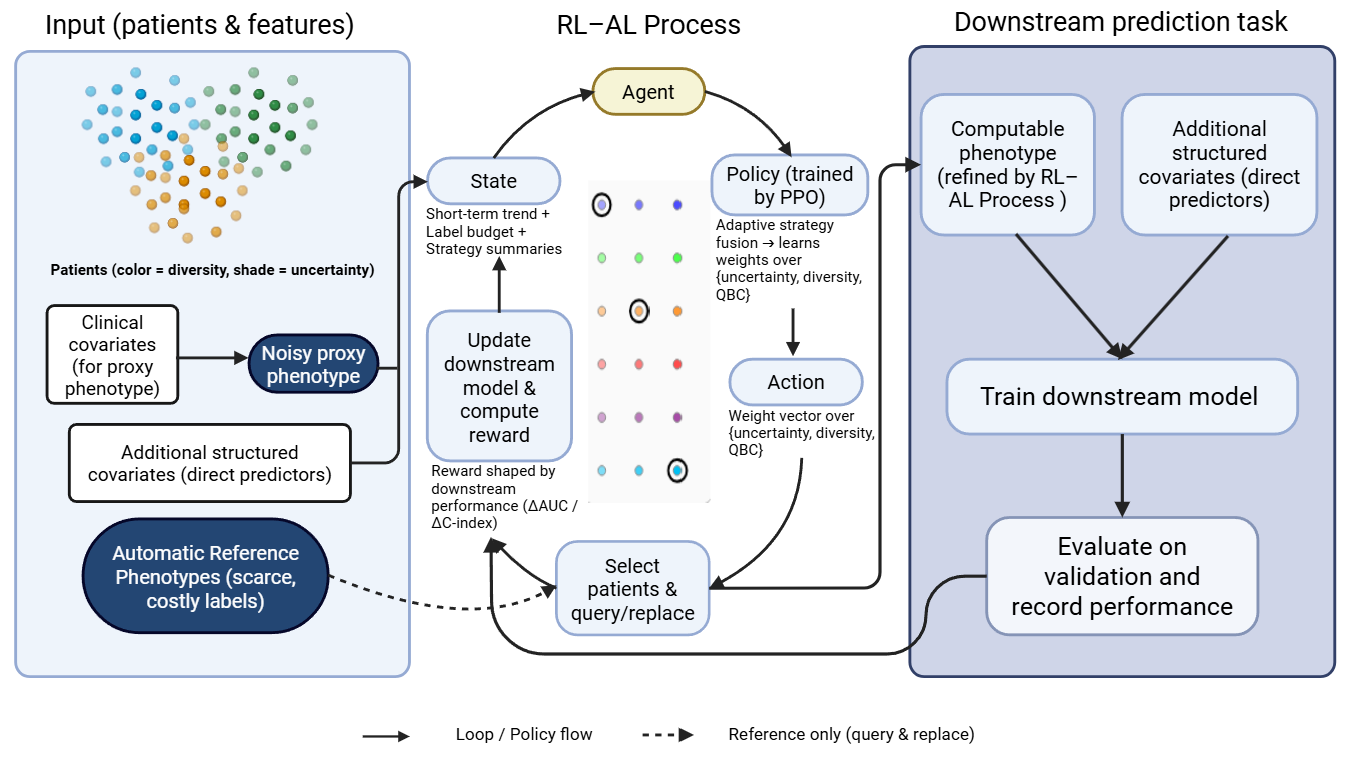}
    \caption{
        RELEAP workflow. Patient-level inputs include clinical covariates for constructing proxy phenotypes, additional structured predictors for risk models, and a limited pool of costly reference labels. 
        The reinforcement learning-active learning (RL-AL) process adaptively selects patients to correct noisy proxy phenotypes into refined computable phenotypes, guided by a policy trained with proximal policy optimization (PPO). 
        The state vector summarizes short-term downstream performance trends, labeling budget, and strategy-level statistics. 
        The policy outputs weights across active learning strategies (uncertainty, diversity, and query-by-committee, QBC), which drive the action of selecting patients for relabeling. 
        Corrected phenotypes are then combined with additional structured predictors to train downstream risk prediction models. 
        Model performance on validation data (e.g., area under the receiver operating characteristic curve [AUC] or concordance index [C-index]) is logged as the final outcome and also fed back as the reward signal, closing the loop between downstream prediction and the RL-AL process.
    }
    \label{fig:rl_flow}
\end{figure}

The RELEAP workflow (Figure~\ref{fig:rl_flow}) begins with a proxy phenotype ($S^{\ast}$) derived from structured data ($X_1$), anchored by a small seed set labeled with the automatic reference phenotype ($S_{\text{true}}$), constructed by combining structured self-reported smoking status with large language model (LLM)-extracted note evidence prior to the index date. A downstream risk prediction model for incident lung cancer ($Y$; time-to-event $T$) provides feedback throughout the process. 
At each iteration, candidate unlabeled patients are scored using multiple active learning heuristics (uncertainty, diversity, and query-by-committee, QBC), as described in Section~\ref{sec:alrl}. A reinforcement learning (RL) agent then adaptively learns how to weight these strategies to select the most informative samples, with detailed reward formulations provided in Supplement~S4. Throughout training, RELEAP maintains proxy labels ($S^{\ast}$) for all patients; only selected samples have their $S^{\ast}$ replaced with $S_{\text{true}}$ in each iteration. Newly acquired labels overwrite the corresponding proxies, iteratively refining the training set. 
The downstream model is retrained after every iteration, and a shaped reward based on predictive performance is used to update the RL policy. Training episodes terminate when the labeling budget is exhausted. This closed feedback loop allows RELEAP to balance exploration and exploitation,\cite{settles2012active} progressively aligning corrected phenotypes with true patient characteristics and improving predictive accuracy and calibration.

\subsubsection{Aphrodite-based baseline}
As a pragmatic comparator, we implemented the Aphrodite method,\cite{banda2017aphrodite} a noisy-label baseline that relies on structured ICD codes and NLP-derived Concept Unique Identifiers (CUIs) without adaptive querying. 
Positive examples were seeded by the presence of smoking-related ICD codes (e.g., V15.82, 305.1, F17.*, Z72.0, Z87.891, Z71.6) and/or affirmed NLP CUI mentions, whereas negatives required no smoking-related ICD codes, no affirmed CUI mentions, and zero pre-index notes. 
Using these seeds, we trained a regularized logistic regression model on per-code counts (ICD/CUI) augmented with a note-count feature, and then scored all patients. 
This noisy but scalable baseline follows the Aphrodite framework for weakly supervised phenotype labeling in EHR data.

\subsubsection{Active Learning Strategies}
We consider three active learning strategies (uncertainty, diversity, and QBC) as the \emph{action basis} for the RELEAP agent. At each iteration, candidate patients are scored by these strategies, normalized, and then combined by the agent to select the top-$k$ samples. Random sampling is included as a comparator but is not part of the RL action space.  

All features are standardized as $z$-scores computed on the \emph{current labeled training set}. For labeled patients, we use the automatic reference phenotype label $S_{\text{true}}$, whereas for unlabeled patients we rely on the continuous proxy score $S^{\ast}$. To ensure comparability, all strategy scores are min–max normalized to the range $[0,1]$ in each iteration. Ties are resolved by adding a small random jitter ($\epsilon \approx 10^{-6}$). Once selected, patients are labeled without replacement and are never re-queried. 

The strategies capture complementary notions of informativeness:  

\begin{enumerate}
  \item \textbf{Uncertainty.}  
        This strategy prioritizes patients whose outcomes are most ambiguous under the current model. 
        For each unlabeled patient $i$, uncertainty is quantified by the entropy of the predicted probability $\hat p_{t,i}$:  
        \[
        H(\hat p_{t,i}) = -\hat p_{t,i}\log \hat p_{t,i} - (1-\hat p_{t,i})\log\!\big(1-\hat p_{t,i}\big),
        \]
        favoring samples for which the model is least confident.\cite{king_logistic_2001}

  \item \textbf{Diversity.}  
        While uncertainty targets ambiguous cases, diversity promotes coverage across the feature space. 
        Distances are computed between each unlabeled patient and its nearest labeled neighbors, with higher scores assigned to patients farther from the labeled set. 
        A variance adjustment is applied for numerical stability (see Supplement~S2).\cite{sener_active_2018}

  \item \textbf{Query-by-Committee (QBC).}  
        To capture model disagreement, we train a committee of $M = 7$ logistic regression models on bootstrap resamples of the labeled set. 
        Each model applies mild perturbations, including small random jitter in the L2 penalty and feature dropout with probability $0.1$. 
        The QBC score reflects disagreement across committee models, measured by the prediction variance with an entropy-based stabilizer (see Supplement~S2). 
        This strategy favors patients where competing models disagree most.\cite{freund_selective_1997}

  \item \textbf{Random (baseline).}  
        Finally, to provide a reference benchmark, we include random sampling, which selects patients uniformly from the unlabeled pool. 
        This strategy is excluded from the RL-driven action space.
\end{enumerate}

\subsubsection{Reinforcement Learning Integration}\label{sec:alrl}

We model phenotype correction under a label budget as a partially observable Markov decision process (POMDP). At each iteration, the agent observes a summary state, outputs a nonnegative weight vector, receives a shaped reward from the downstream model’s performance, and transitions to the next state. Episodes terminate when the labeling budget is exhausted.

\paragraph{State.}  
The state $s_t$ summarizes the current learning context, including per-strategy score summaries on the unlabeled pool, composition of the labeled set, the recent trajectory of downstream performance, and the remaining budget (see Supplement S3 for full details).  

\paragraph{Action.}  
Based on the observed state, the agent outputs a nonnegative \textbf{weight vector} over the three informative strategies (uncertainty, diversity, and QBC):
\[
\mathbf{w}_t = \big(w^{\text{unc}},\, w^{\text{div}},\, w^{\text{qbc}}\big), \quad w_i \ge 0, \quad \sum_i w_i = 1.
\]
This vector combines the normalized strategy scores into a single ranking function that determines which patients are selected from the unlabeled pool.  

\paragraph{Reward.}
The agent receives a shaped reward derived from the downstream model’s performance (e.g., AUC in classification or C-index in survival). Stabilization is achieved via a short moving baseline and mild budget-aware scaling; detailed formulas are provided in Supplement S3.  

\paragraph{RL algorithm.}  
The policy $\pi_\theta(a \mid s)$ is optimized with Proximal Policy Optimization (PPO), maximizing the expected discounted return $\mathbb{E}[\sum_{t=1}^{T}\gamma^{t-1}R_t]$ (see Supplement S3).  

\paragraph{Iteration loop.}  
Each iteration follows an agent-environment cycle: the agent observes $s_t$, selects action $\mathbf{w}t$, receives reward $r_t$, and transitions to $s{t+1}$.
The practical implementation proceeds as follows:
\begin{enumerate}
  \item Compute strategy scores on the unlabeled pool using uncertainty, diversity, and QBC (with random sampling used only as a baseline).
  \item Observe the current state $s_t$, including strategy summaries, recent trajectory, budget fraction, labeled-set composition, and downstream performance.
  \item Select an action $\mathbf{w}_t$ via the PPO policy and rank the unlabeled pool accordingly.
  \item Acquire labels by querying $S_{\text{true}}$ for the top-$k$ selected patients, overwriting $S^{\ast}$ with $S_{\text{true}}$, and marking them as labeled. After this update, the phenotype field $S$ refers to the full training set, where labeled patients use $S_{\text{true}}$ and all remaining patients continue to contribute through their $S^{\ast}$ values.
  \item Retrain the downstream prediction model (logistic or penalized Cox, depending on mode) and evaluate validation performance; full details on feature processing and additional metrics are in Supplement S3.
  \item Compute the reward $R_t$, store $(s_t, a_t, R_t, s_{t+1})$, and update the PPO agent periodically.  
\end{enumerate}
Details of the simulation setup and synthetic data experiments are provided in Supplement~S1.

\subsection{Evaluation of RELEAP on Real-World EHR Data}

We evaluated RELEAP using de-identified EHR data from the DUHS spanning \textbf{2014--2024}. This dataset provides a representative testbed for assessing phenotype correction in the context of lung cancer risk prediction, where accurate characterization of smoking behavior is particularly critical. As noted earlier, ICD-only proxies ($S^{\ast}$) tend to underreport smoking compared to the multimodal automatic reference phenotype ($S_{\text{true}}$), motivating evaluation in this setting.  

\paragraph{Cohort and study population.}  
The study cohort included adults aged 35–70 years with at least 365 days of prior EHR history before the index date, defined as the qualifying outpatient encounter. Duplicate patient records were removed. Patients with evidence of prevalent lung cancer prior to the index date were excluded. Records with missing demographic information (e.g., sex, race) were retained and handled using predefined preprocessing rules.  

\paragraph{Variable construction and notation.}  
Study variables were derived from both structured and unstructured EHR elements. Smoking-related ICD-9/10 codes were extracted within a one-year look-back window prior to the index date and encoded as features ($X_1$). A proxy phenotype ($S^{\ast}$) was estimated using a logistic regression model trained on $X_1$.  
The automatic reference phenotype ($S_{\text{true}}$) was defined by combining structured self-reported smoking status with LLM-extracted mentions from clinical notes prior to index, prioritizing structured entries in case of disagreement.  
Additional structured covariates ($X_2$) included demographics, chronic obstructive pulmonary disease (COPD; binary indicator and one-year pre-index count), and a Hispanic ethnicity indicator, all measured at baseline.  
The outcome ($Y$) was incident lung cancer after the index date, defined as $\geq$2 primary lung cancer ICD codes occurring within 60 days of each other before the administrative censoring date.  
Time-to-event ($T$) was measured from the index date to the first qualifying code; patients without an event were censored at the truncation date.  

\paragraph{Automatic reference phenotype definitions and variants.}  
\label{sec:arp-definitions}
To assess robustness, we considered several definitions of the automatic reference phenotype.  
Unless otherwise noted, the primary definition integrates structured self-reported smoking status with smoking-related mentions extracted from clinical notes using large language models.  
For sensitivity analyses, we also evaluated three alternative definitions:  
(1) combining structured self-reported smoking status with concept identifiers extracted by traditional NLP methods;  
(2) using only concept identifiers extracted by traditional NLP without relying on self-reported information; and  
(3) using only smoking-related mentions identified by LLMs without self-reported data.  
Comparative results for these variants are summarized in Supplement~S5.

\begin{figure}[h]
    \centering
    \includegraphics[scale=0.6]{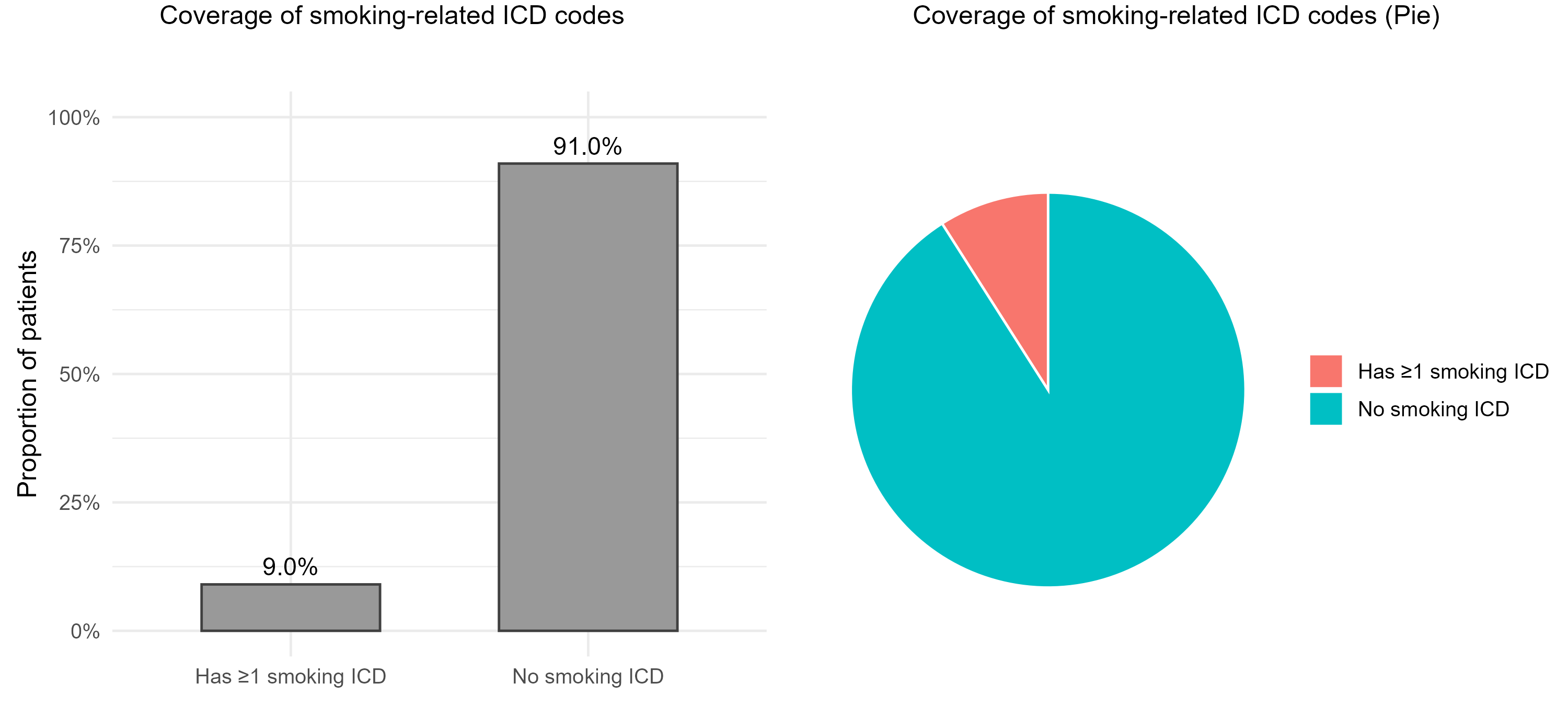}
  \caption{Coverage of smoking-related ICD codes (presence of $\geq 1$ smoking-related ICD per patient).}
  \label{fig:icd-coverage}
\end{figure}

\paragraph{Preprocessing.}  
Categorical variables were one-hot encoded with an explicit “Unknown” category for missing values. Continuous variables were standardized, and missing values were imputed using the training-set median.  
All features and $S_{\text{true}}$ were constructed strictly from pre-index data to prevent data leakage.  
A descriptive summary of the final cohort, including demographic characteristics, smoking phenotype distributions (based on $S_{\text{true}}$), outcome rates, COPD prevalence, and coverage of smoking-related ICD codes, is provided in Table~\ref{tab:dataset-summary}.  
Figure~\ref{fig:icd-coverage} further illustrates the proportion of patients with at least one smoking-related ICD code.

\captionsetup{width=.95\textwidth}

\renewcommand{\arraystretch}{0.8}

\begin{longtable}{@{}p{0.55\linewidth}>{\raggedleft\arraybackslash}p{0.4\linewidth}@{}}
\caption{Dataset summary of the DUHS lung cancer cohort.}
\label{tab:dataset-summary}\\
\hline
\textbf{Characteristic} & \textbf{Value} \\
\hline
\endfirsthead

\hline
\textbf{Characteristic} & \textbf{Value} \\
\hline
\endhead

\hline
\multicolumn{2}{r}{\small\emph{(continued on next page)}}\\
\hline
\endfoot

\hline
\endlastfoot

\textbf{Overall} & \\
\hspace{1em} N & 238,119 \\
\hspace{1em} Age, mean (SD) & 52.94 (10.23) \\
\hspace{1em} Age, median [IQR] & 53 [44--62] \\

\textbf{Sex} & \\
\hspace{1em} Female & 138,415 (58.1\%) \\
\hspace{1em} Male & 99,702 (41.9\%) \\
\hspace{1em} No information & 1 (0.0\%) \\
\hspace{1em} Unknown & 1 (0.0\%) \\

\textbf{Race} & \\
\hspace{1em} American Indian or Alaska Native & 1,512 (0.6\%) \\
\hspace{1em} Asian & 8,357 (3.5\%) \\
\hspace{1em} Black or African American & 57,014 (23.9\%) \\
\hspace{1em} Multiple race & 926 (0.4\%) \\
\hspace{1em} Native Hawaiian or Other Pacific Islander & 227 (0.1\%) \\
\hspace{1em} No information & 318 (0.1\%) \\
\hspace{1em} Other & 4,184 (1.8\%) \\
\hspace{1em} Refuse to answer & 6,227 (2.6\%) \\
\hspace{1em} White & 159,354 (66.9\%) \\

\textbf{Hispanic/Latino} & \\
\hspace{1em} Declined to report & 8,844 (3.7\%) \\
\hspace{1em} No & 219,828 (92.3\%) \\
\hspace{1em} Unavailable & 187 (0.1\%) \\
\hspace{1em} Unknown & 24 (0.0\%) \\
\hspace{1em} Yes & 9,236 (3.9\%) \\

\textbf{Smoking} & \\
\hspace{1em} Never smoked & 125,556 (52.7\%) \\
\hspace{1em} No information & 14,940 (6.3\%) \\
\hspace{1em} Smoked in the past & 50,301 (21.1\%) \\
\hspace{1em} Smokes daily & 19,717 (8.3\%) \\
\hspace{1em} Smokes daily (heavy) & 204 (0.1\%) \\
\hspace{1em} Smokes daily (light) & 1,345 (0.6\%) \\
\hspace{1em} Smokes some days & 3,863 (1.6\%) \\
\hspace{1em} Smoking status unknown & 89 (0.0\%) \\
\hspace{1em} Unknown & 22,104 (9.3\%) \\

\textbf{Outcome} & \\
\hspace{1em} Lung cancer, n (\%) & 1,846 (0.78\%) \\

\textbf{Smoking ICD coverage} & \\
\hspace{1em} No smoking ICD & 216,614 (91.0\%) \\
\hspace{1em} Has $\geq$ 1 smoking ICD & 21,505 (9.0\%) \\

\textbf{COPD} & \\
\hspace{1em} No & 226,055 (94.9\%) \\
\hspace{1em} Yes & 12,064 (5.1\%) \\

\end{longtable}

\vspace{-1em}
\noindent\footnotesize\textbf{Notes.}
\textit{SEX} values are expanded to full names;
\textit{RACE} uses single-code mapping (01--07, NI, UN, OT);
\textit{HISPANIC} is reported separately (Y, N, R, UN, NI).
\textit{Smoking behavior} was derived from vitals labels in column \texttt{smoking\_status} and presented as behavioral descriptors
(Smokes daily / Smokes some days / Smoked in the past / Never smoked / Unknown).
\textit{Heavy} and \textit{light} correspond to vitals labels (e.g., `heavy tobacco smoker' and `light tobacco smoker'); no quantitative cigarettes-per-day definition was available in this dataset.
\textit{Smoking ICD coverage} is defined as having $\geq 1$ smoking-related ICD codes.
\normalsize

\paragraph{Experimental setup.}  
Experiments followed a training-validation framework to simulate real-world phenotype correction. Patients were randomly split into 80\% training and 20\% validation subsets, stratified by the outcome to preserve event rates. Active learning began with a small, balanced seed set drawn from the training partition, where the noisy proxy labels ($S^{\ast}$) were replaced by the automatic reference labels ($S_{\text{true}}$) to initialize the labeled set.  
All remaining training patients continued to contribute through their proxy labels, ensuring that the entire training pool was used for model updates.  
At each iteration, the RELEAP agent generated a set of weights across candidate sampling strategies to guide patient selection.  
For evaluation, the same weighting scheme was mirrored on the validation set so that training and validation evolved under comparable replacement dynamics.  
Downstream prediction models were retrained at every iteration, and their predictive performance was tracked in parallel.

\paragraph{Planned analyses.}  
The primary classification task predicted incident lung cancer using the combined feature set of the phenotype estimate ($S$) and structured covariates ($X_2$).  
Model performance was primarily evaluated by the AUC.  
A secondary analysis used a penalized Cox proportional hazards model to estimate time-to-event outcomes, with predictive accuracy summarized by the C-index.  
Performance across iterations was summarized as the mean value with 95\% confidence intervals across repeated simulations.  
Final-round comparisons included measures of discrimination (AUC or C-index), calibration (mean squared error [MSE] of predicted probabilities), and threshold-based metrics such as the F1 score, true positive rate (TPR), and positive predictive value (PPV) at a fixed false positive rate (FPR) of 0.1.  
We also conducted subgroup analyses stratified by sex, not for clinical inference but to illustrate how different active learning strategies can behave differently under varying data distributions.

\paragraph{Comparators.}
We benchmarked RELEAP against a hierarchy of baseline models:

\begin{enumerate}[label=(\roman*), leftmargin=*, itemsep=2pt, topsep=2pt]
  \item \textbf{Proxy-only} (\texttt{S\_star\_baseline}): a zero-label baseline trained directly on the noisy proxy phenotype ($S^{\ast}$) without label correction.
  \item \textbf{Aphrodite-based baseline} (\texttt{Aphrodite}): a weakly supervised method relying on structured ICD codes and NLP-derived CUIs without adaptive querying.
  \item \textbf{Single-strategy active learning}: \texttt{random}, \texttt{uncertainty}, \texttt{diversity}, \texttt{qbc}.
  \item \textbf{RELEAP}: a performance-driven fusion of multiple sampling strategies.
  \item \textbf{Oracle} (\texttt{S\_true\_oracle}): a fully supervised model trained with all high-fidelity automatic reference phenotype labels ($S_{\text{true}}$).
\end{enumerate}

For completeness, we also compared RELEAP across multiple automatic reference phenotype definitions described in Section~2.3 and summarized results in Supplement~S5.

\section{RESULTS}

We evaluated RELEAP on the DUHS lung cancer cohort to assess its effectiveness in correcting noisy proxy phenotypes.  
Results are organized into two parts: (1) overall performance across all patients, considering both logistic classification and survival analyses, and (2) subgroup analyses stratified by sex to illustrate how different active learning strategies behave under varying data distributions.

\subsection{Overall Performance Across All Patients}

RELEAP was evaluated under two settings, logistic classification and survival analysis.  
In both modes, RELEAP substantially improved predictive performance over the noisy proxy phenotype baseline ($S^{\ast}$) and approached the oracle trained with all automatic reference phenotype labels ($S_{\text{true}}$).  
Sensitivity analyses (Supplement~S5) showed that the combined definition using structured self-reports and LLM-extracted notes achieved higher AUC and lower MSE than the NLP-based variant, and both SR-based definitions outperformed their NoSR counterparts.

\begin{figure}[h]
    \centering
    \includegraphics[scale=0.32]{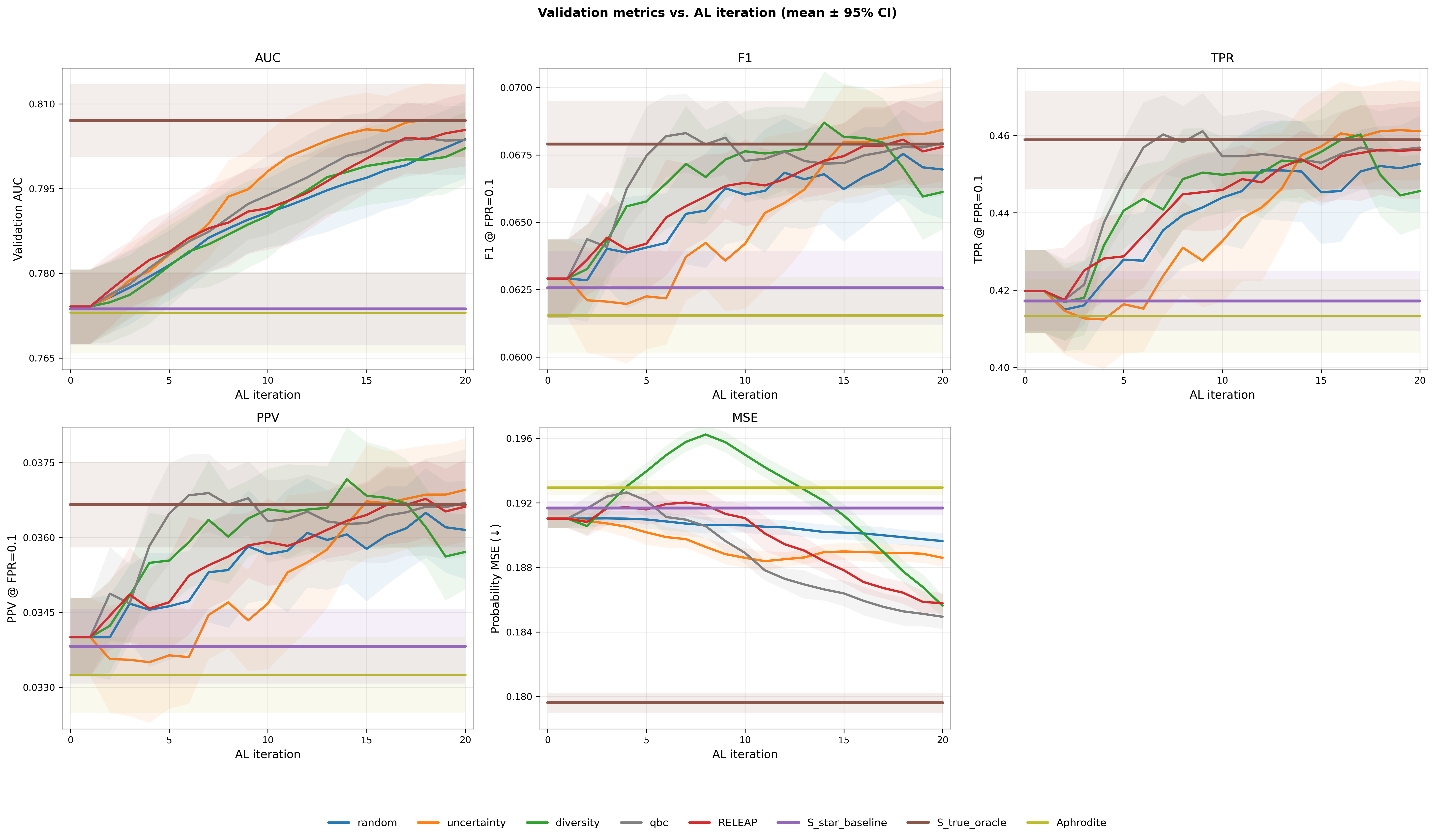}
    \caption{Validation Metrics vs. AL Iteration in the Logistic Mode (mean $\pm$ 95\% CI over ten replications).}
    \label{fig:al_facets}
\end{figure}

\paragraph{Logistic classification mode (evaluated by AUC).}  
Models were trained on labeled $[S, X_2]$ features and evaluated on the held-out validation set.  
Each experiment used a total labeling budget of 63{,}000 patients, queried in batches of 3{,}000 across 20 iterations, with an initial balanced seed set of 3{,}000.  
All methods were repeated over ten independent replications.  
Figure~\ref{fig:al_facets} shows mean $\pm$ 95\% confidence interval trajectories for discrimination (AUC, F1 score, TPR, PPV) and calibration (probability MSE).

At convergence, discrimination improved markedly: mean AUC rose from $0.774 \pm 0.010$ for the noisy proxy baseline to $0.805$ with RELEAP and $0.807$ with uncertainty sampling, closely approaching oracle-level performance ($0.807$).  
F1 and TPR showed comparable gains, while PPV increased modestly from $0.034$ to $0.037$.  
Calibration also improved, with MSE decreasing from $0.192$ (baseline) to $0.186$ under RELEAP, approaching the oracle’s $0.180$.  

Strategy behaviors diverged across metrics.  
\textbf{RELEAP} consistently ranked among the top performers with smoother, more stable trajectories.  
\emph{Uncertainty} sampling achieved the highest short-term gains but with greater variability.  
\emph{Diversity} favored calibration but lagged slightly in discrimination, while \emph{QBC} improved F1 score early before stabilizing.  
\emph{Random} sampling improved gradually but remained inferior overall.  
RELEAP’s advantage lay in adaptively balancing these complementary heuristics.

\begin{figure}[h]
    \centering
    \includegraphics[scale=0.6]{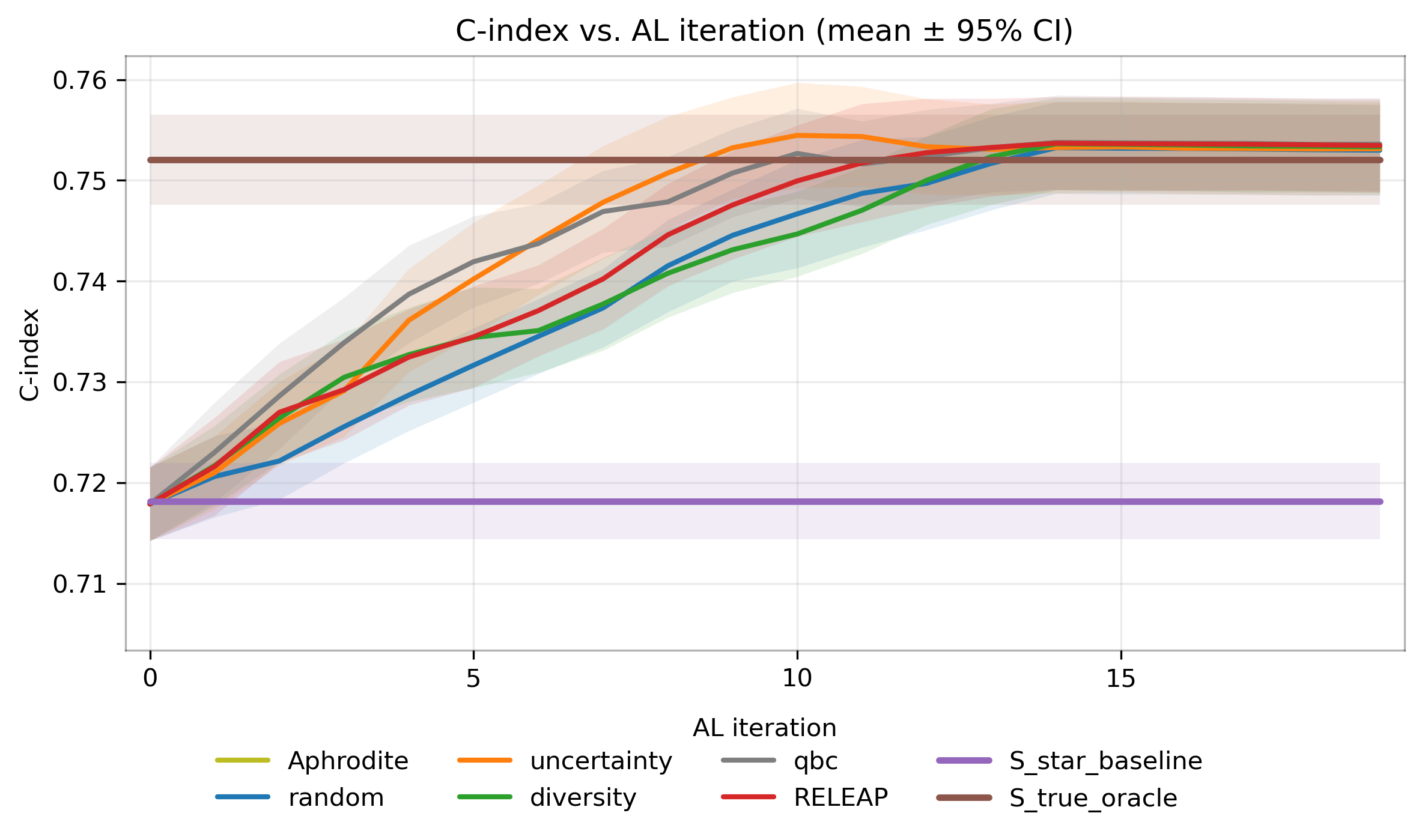}
    \caption{C-index vs. AL Iteration in the Survival (Cox) Mode (mean $\pm$ 95\% CI over ten replications).}
    \label{fig:cindex_curve}
\end{figure}

\paragraph{Survival mode (evaluated by C-index).}  
In the survival setting (penalized Cox regression; $Z$ constructed from $S$ and $X_2$ with feature screening), experiments used a total labeling budget of 84{,}000 patients, queried in batches of 4{,}000 over 20 iterations, with an initial balanced seed set of 4{,}000.  
Each method was repeated ten times.  
Figure~\ref{fig:cindex_curve} demonstrates that all strategies markedly outperformed the noisy proxy baseline (C-index $0.718$, 95\% CI: 0.713--0.723).  
RELEAP achieved $0.752$ (0.747--0.757), converging toward the oracle ($0.753$, 0.748--0.758).  

\emph{Uncertainty} and \emph{QBC} reached this plateau fastest, followed closely by \textbf{RELEAP}, while \emph{Diversity} and \emph{Random} trailed slightly.  

Overall, RELEAP provided the most balanced and reliable trajectory, achieving robust survival prediction with convergence, mirroring the efficiency observed in the logistic classification experiments.

\subsection{Subgroup Analyses: Male vs.\ Female}

\begin{figure}[!htbp]
    \centering
    \includegraphics[scale=0.3]{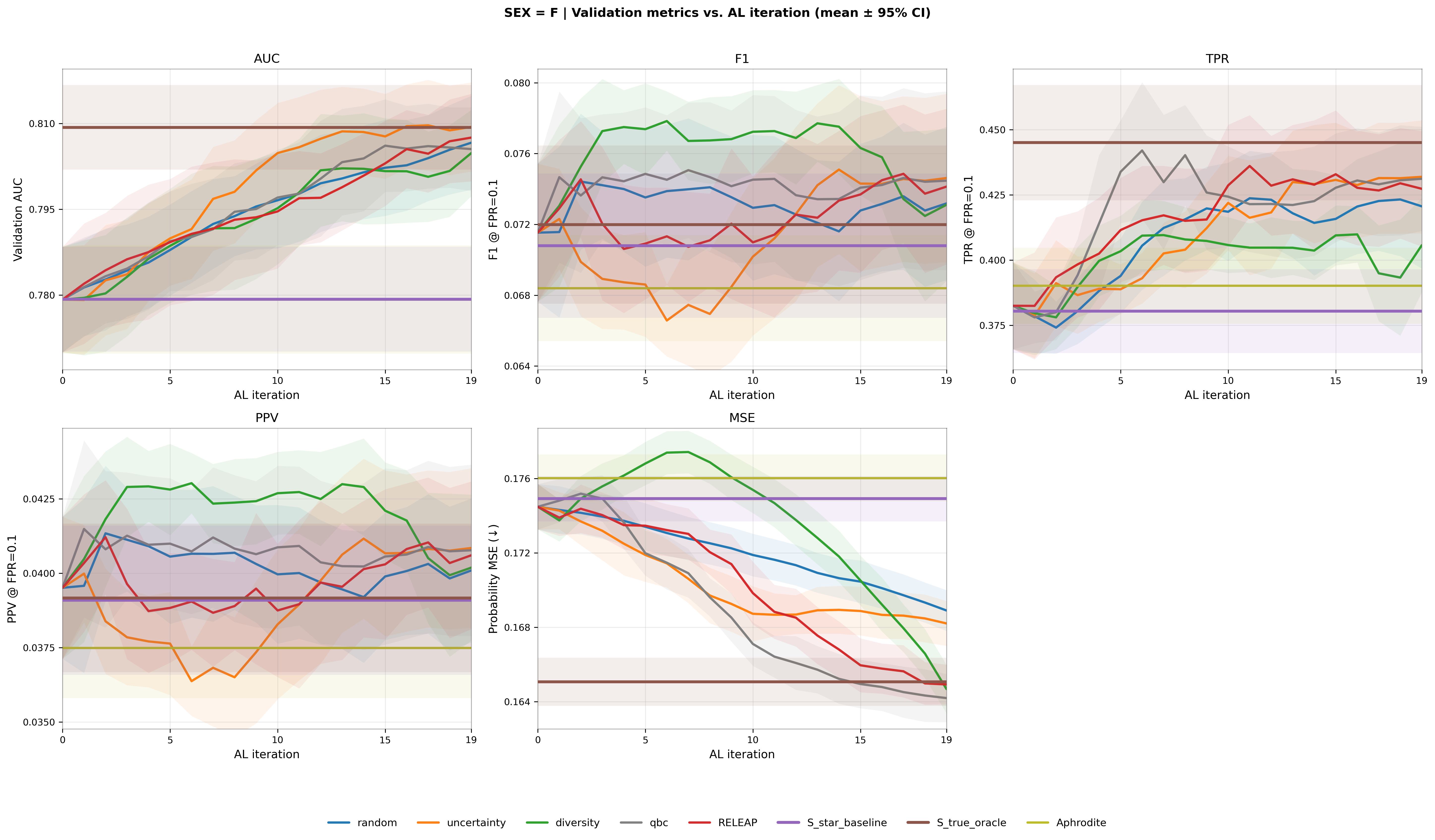}
    \includegraphics[scale=0.3]{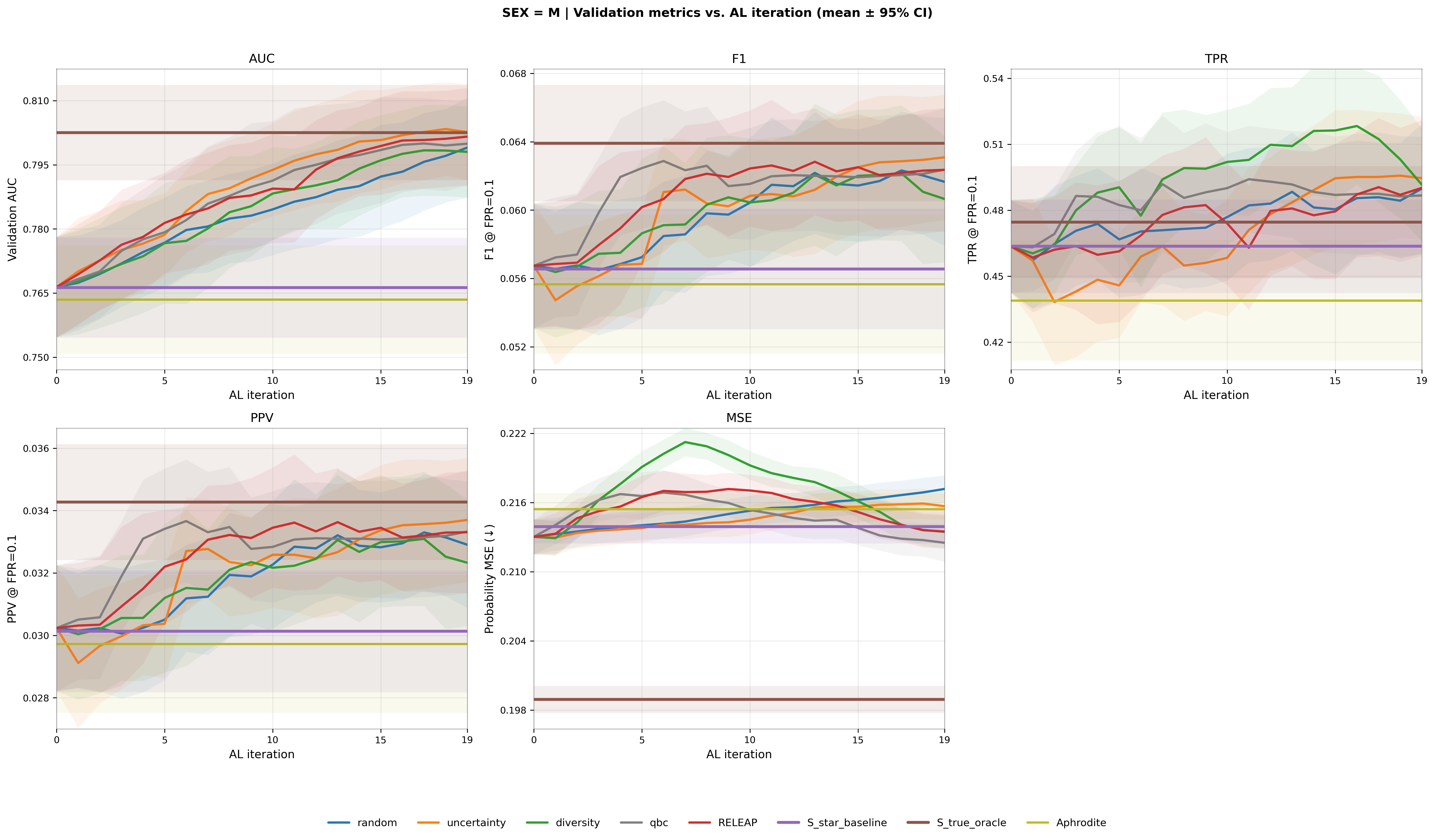}
    \caption{Validation Metrics vs.\ AL Iteration in the Logistic Mode (mean $\pm$ 95\% CI over ten replications) within male and female subgroups.}
    \label{fig:al_facets_subgroups}
\end{figure}

To further examine population heterogeneity, we stratified the analyses by sex.  
Figure~\ref{fig:al_facets_subgroups} presents validation trajectories (mean $\pm$ 95\% CI over ten replications) for AUC, F1 score, TPR, PPV, and probability MSE in male and female subgroups.

\paragraph{Male subgroup.}  
All active learning strategies improved steadily over the noisy proxy baseline (AUC $0.766 \pm 0.019$) and approached the automatic reference phenotype oracle ($0.803 \pm 0.018$).  
Final AUC values clustered around 0.798–0.803 across methods, with RELEAP ($0.802 \pm 0.018$) and uncertainty sampling ($0.803 \pm 0.018$) leading.  
F1 increased modestly from $0.057 \pm 0.006$ to about 0.063, with RELEAP and QBC showing smoother convergence.  
TPR rose from $0.464 \pm 0.034$ to nearly 0.495 (uncertainty), while PPV improved slightly from 0.030 to 0.033.  
Calibration gains were limited: MSE decreased marginally from $0.214 \pm 0.002$ to about 0.213 under RELEAP and QBC, remaining above the oracle ($0.199 \pm 0.002$).  
Overall, all strategies converged to similar discrimination and calibration levels, with RELEAP offering smoother and more stable trajectories rather than absolute superiority.

\paragraph{Female subgroup.}  
In contrast, female subgroup results revealed clearer separation across strategies.  
Baseline AUC ($0.779 \pm 0.015$) improved to around 0.808–0.809 under RELEAP ($0.808 \pm 0.012$), uncertainty ($0.809 \pm 0.013$), and QBC ($0.806 \pm 0.012$), closely matching the oracle ($0.809 \pm 0.012$).  
F1 increased from $0.071 \pm 0.007$ to 0.074–0.075, with RELEAP and QBC showing smoother trajectories, while diversity exhibited sharper fluctuations.  
TPR rose from $0.380 \pm 0.026$ to above 0.430 for QBC, uncertainty, and RELEAP, approaching the oracle ($0.445 \pm 0.036$).  
PPV improved from $0.039 \pm 0.004$ to approximately 0.041 under RELEAP, QBC, and uncertainty.  
Calibration gains were most pronounced in females: MSE decreased from $0.175 \pm 0.002$ to near-oracle levels ($0.165 \pm 0.002$) under RELEAP, QBC, and diversity, while random and uncertainty plateaued slightly higher ($\approx0.168$).

\paragraph{Summary.}  
In summary, while male subgroup results showed comparable convergence across strategies with limited calibration improvement, female subgroup results demonstrated clearer benefits from RELEAP.  
The RELEAP agent achieved balanced gains across discrimination, threshold, and calibration metrics, with the most stable convergence dynamics.  

\FloatBarrier
\section{DISCUSSION}

In this work, we proposed RELEAP, a reinforcement learning-enhanced agent for label-efficient active phenotyping. Unlike traditional active learning approaches that rely on a fixed querying strategy, RELEAP dynamically adjusts the weighting of multiple heuristics based on downstream model performance. By integrating uncertainty, diversity, and committee-based strategies within an RL-guided policy, RELEAP improves the quality of acquired labels and enhances predictive performance under constrained labeling budgets. The use of reinforcement learning further provides natural flexibility to incorporate multi-metric rewards, enabling balanced improvements in both discrimination and calibration.

Our real-world evaluation using the DUHS EHR cohort demonstrated consistent performance gains. Across both logistic classification and survival analyses, RELEAP effectively closed the gap between noisy proxy phenotypes and automatic reference phenotype labels, achieving near-oracle discrimination while also improving threshold-level metrics and calibration. Subgroup analyses further illustrated that while all strategies performed similarly in stable settings (e.g., males), the RL-guided approach provided clear advantages in groups with greater variability (e.g., females), underscoring the value of adaptivity over fixed heuristics. These findings suggest that reinforcement learning can play an important role in guiding label acquisition, especially in clinical domains where proxies are noisy and manual labeling is costly. By leveraging downstream performance as feedback, RELEAP avoids the limitations of single-strategy active learning and demonstrates robust improvements across settings.

This study has several limitations. First, RELEAP was validated only in the context of smoking phenotype correction for lung cancer risk prediction. While smoking is a salient and clinically important phenotype, further evaluation is needed to assess generalizability across other phenotyping tasks and disease contexts. Second, the current implementation focused on EHR data from a single health system, which may limit external validity. Finally, we considered only three active learning strategies; additional heuristics or feature modalities may further enrich the policy space.  

Future work should extend RELEAP to a broader set of phenotype correction tasks, including behavioral, genomic, and imaging-derived phenotypes. Beyond phenotyping, the agent could be applied to downstream applications such as risk stratification, treatment effect heterogeneity, and trial eligibility screening, where label efficiency and adaptivity are equally important. Incorporating richer reward functions that balance fairness, interpretability, and cost-effectiveness may also further enhance RELEAP’s impact in real-world clinical decision support.

\section{CONCLUSION}
In conclusion, RELEAP provides a practical framework for adaptive, label-efficient phenotyping. By using downstream model performance as a feedback signal, RELEAP dynamically refines phenotype labels to enhance risk prediction accuracy and model reliability in real-world EHR settings.

\section*{COMPETING INTERESTS}
None declared.

\section*{DATA AVAILABILITY STATEMENT}
Simulation code and data are available upon request. EHR data contain protected health information and are not able to be shared.

\printbibliography

@article{hong2019semi,
  title={Semi-supervised validation of multiple surrogate outcomes with application to electronic medical records phenotyping},
  author={Hong, Chuan and Liao, Katherine P and Cai, Tianxi},
  journal={\textit{Biometrics}},
  volume={75},
  number={1},
  pages={78--89},
  year={2019},
  publisher={Wiley Online Library}
}

@article{hong2023international,
  title={International classification of diseases (ICD)},
  author={Hong, Yi and Zeng, Marcia Lei},
  journal={\textit{KO Knowledge Organization}},
  volume={49},
  number={7},
  pages={496--528},
  year={2023},
  publisher={Nomos Verlagsgesellschaft mbH \& Co. KG}
}

@article{ruckdeschel2023unstructured,
  title={Unstructured data are superior to structured data for eliciting quantitative smoking history from the electronic health record},
  author={Ruckdeschel, John C and Riley, Mark and Parsatharathy, Sriram and Chamarthi, Rajesh and Rajagopal, Chakethraman and Hsu, Hui Shuang and Mangold, Doug and Driscoll, Chiny},
  journal={\textit{JCO Clin Cancer Inform}},
  volume={7},
  pages={e2200155},
  year={2023},
  publisher={Wolters Kluwer Health}
}

@article{chen2013applying,
  title={Applying active learning to high-throughput phenotyping algorithms for electronic health records data},
  author={Chen, Yukun and Carroll, Robert J and Hinz, Eugenia R McPeek and Shah, Anushi and Eyler, Anne E and Denny, Joshua C and Xu, Hua},
  journal={\textit{J Am Med Inform Assoc}},
  volume={20},
  number={e2},
  pages={e253--e259},
  year={2013},
  publisher={BMJ Publishing Group BMA House, Tavistock Square, London, WC1H 9JR}
}

@article{nogues2022weakly,
  title={Weakly semi-supervised phenotyping using electronic health records},
  author={Nogues, Isabelle-Emmanuella and Wen, Jun and Lin, Yucong and Liu, Molei and Tedeschi, Sara K and Geva, Alon and Cai, Tianxi and Hong, Chuan},
  journal={\textit{J Biomed Inform}},
  volume={134},
  pages={104175},
  year={2022},
  publisher={Elsevier}
}

@article{hong2025label,
  title={Label efficient phenotyping for Long COVID using electronic health records},
  author={Hong, Chuan and Wen, Jun and Zhang, Harrison G and Ayakulangara Panickan, Vidul and Yang, Doris Y and Chen, Alicia W and Xiong, Xin and Wang, Xuan and Morris, Michele and Morini, Sara and others},
  journal={\textit{npj Digit Med}},
  volume={8},
  number={1},
  pages={405},
  year={2025},
  publisher={Nature Publishing Group UK London}
}

@article{liao2019high,
  title={High-throughput multimodal automated phenotyping (MAP) with application to PheWAS},
  author={Liao, Katherine P and Sun, Jiehuan and Cai, Tianrun A and Link, Nicholas and Hong, Chuan and Huang, Jie and Huffman, Jennifer E and Gronsbell, Jessica and Zhang, Yichi and Ho, Yuk-Lam and others},
  journal={\textit{J Am Med Inform Assoc}},
  volume={26},
  number={11},
  pages={1255--1262},
  year={2019},
  publisher={Oxford University Press}
}

@article{denny2010phewas,
  title={PheWAS: demonstrating the feasibility of a phenome-wide scan to discover gene--disease associations},
  author={Denny, Joshua C and Ritchie, Marylyn D and Basford, Melissa A and Pulley, Jill M and Bastarache, Lisa and Brown-Gentry, Kristin and Wang, Deede and Masys, Dan R and Roden, Dan M and Crawford, Dana C},
  journal={\textit{Bioinformatics}},
  volume={26},
  number={9},
  pages={1205--1210},
  year={2010},
  publisher={Oxford University Press}
}

@article{zhang2019high,
  title={High-throughput phenotyping with electronic medical record data using a common semi-supervised approach (PheCAP)},
  author={Zhang, Yichi and Cai, Tianrun and Yu, Sheng and Cho, Kelly and Hong, Chuan and Sun, Jiehuan and Huang, Jie and Ho, Yuk-Lam and Ananthakrishnan, Ashwin N and Xia, Zongqi and others},
  journal={\textit{Nat Protoc}},
  volume={14},
  number={12},
  pages={3426--3444},
  year={2019},
  publisher={Nature Publishing Group UK London}
}

@article{chandran2023machine,
  title={Machine learning and real-world data to predict lung cancer risk in routine care},
  author={Chandran, Urmila and Reps, Jenna and Yang, Robert and Vachani, Anil and Maldonado, Fabien and Kalsekar, Iftekhar},
  journal={\textit{Cancer Epidemiol Biomarkers Prev}},
  volume={32},
  number={3},
  pages={337--343},
  year={2023},
  publisher={American Association for Cancer Research}
}

@article{banda2017aphrodite,
  title = {Electronic phenotyping with APHRODITE and the Observational Health Sciences and Informatics (OHDSI) data network},
  author = {Banda, Juan M. and Halpern, Yoni and Sontag, David and Shah, Nigam H.},
  journal = {\textit{AMIA Jt Summits Transl Sci Proc}},
  volume = {2017},
  pages = {48--57},
  year = {2017},
  pmid = {28815104},
  pmcid = {PMC5543379},
  url = {https://pubmed.ncbi.nlm.nih.gov/28815104/}
}

@book{settles2012active,
  title={Active Learning},
  author={Settles, Burr},
  publisher={Morgan \& Claypool},
  year={2012},
  series={Synthesis Lectures on Artificial Intelligence and Machine Learning},
  doi={10.2200/S00429ED1V01Y201207AIM018}
}

@article{shivade_review_2014,
  title = {A review of approaches to identifying patient phenotype cohorts using electronic health records},
  volume = {21},
  number = {2},
  issn = {1527-974X},
  url = {https://pmc.ncbi.nlm.nih.gov/articles/PMC3932460/},
  doi = {10.1136/amiajnl-2013-001935},
  pages = {221--230},
  journal = {\textit{J Am Med Inform Assoc}},
  shortjournal = {J Am Med Inform Assoc},
  author = {Shivade, Chaitanya and Raghavan, Preethi and Fosler-Lussier, Eric and Embi, Peter J and Elhadad, Noemie and Johnson, Stephen B and Lai, Albert M},
  urldate = {2025-09-21},
  date = {2013-11},
  langid = {english},
}

@article{halpern_anchor_2016,
  title = {Electronic medical record phenotyping using the anchor and learn framework},
  volume = {23},
  issn = {1527-974X},
  url = {https://pubmed.ncbi.nlm.nih.gov/27107443/},
  doi = {10.1093/jamia/ocw011},
  pages = {731--740},
  journal = {\textit{J Am Med Inform Assoc}},
  shortjournal = {J Am Med Inform Assoc},
  author = {Halpern, Yoni and Horng, Steven and Choi, Youngduck and Sontag, David},
  urldate = {2025-09-21},
  date = {2016-07},
  langid = {english},
}

@article{wiley_icd9_2013,
title = {ICD-9 tobacco use codes are effective identifiers of smoking status},
volume = {20},
issn = {1527-974X},
url = {https://pmc.ncbi.nlm.nih.gov/articles/PMC3721171/}
,
doi = {10.1136/amiajnl-2012-001557},
pages = {652--658},
journal = {\textit{J Am Med Inform Assoc}},
shortjournal = {J Am Med Inform Assoc},
author = {Wiley, Laura K. and Shah, Anushi and Xu, Hua and Bush, William S.},
urldate = {2025-09-21},
date = {2013-07},
langid = {english},
}

@article{rajendran_nlp_smoking_2020,
title = {Extracting Smoking Status from Electronic Health Records Using {NLP} and Deep Learning},
volume = {2020},
issn = {2153-4063},
url = {https://pmc.ncbi.nlm.nih.gov/articles/PMC7233082/},
pages = {507--516},
journal = {\textit{AMIA Jt Summits Transl Sci Proc}},
shortjournal = {AMIA Jt Summits Transl Sci Proc},
author = {Rajendran, Suraj and Topaloglu, Umit},
urldate = {2025-09-21},
date = {2020-05},
langid = {english},
}

@inproceedings{fang_learning_2017,
title = {Learning how to Active Learn: A Deep Reinforcement Learning Approach},
url = {https://aclanthology.org/D17-1063/}
,
doi = {10.18653/v1/D17-1063},
pages = {595--605},
booktitle = {Proceedings of the 2017 Conference on Empirical Methods in Natural Language Processing},
author = {Fang, Meng and Li, Yuan and Cohn, Trevor},
urldate = {2025-09-21},
date = {2017-09},
langid = {english},
}

@article{usptf_lungcancer_2021,
  title = {Screening for Lung Cancer: {US} Preventive Services Task Force Recommendation Statement},
  volume = {325},
  number = {10},
  issn = {0098-7484},
  url = {https://jamanetwork.com/journals/jama/fullarticle/2777243},
  doi = {10.1001/jama.2021.1117},
  pages = {962--970},
  journal = {\textit{JAMA}},
  shortjournal = {JAMA},
  author = {{US Preventive Services Task Force}},
  urldate = {2025-09-21},
  date = {2021-03},
  langid = {english},
}

@article{king_logistic_2001,
  title = {Logistic Regression in Rare Events Data},
  volume = {9},
  number = {2},
  pages = {137--163},
  issn = {1047-1987},
  url = {https://gking.harvard.edu/files/0s.pdf},
  doi = {10.1093/oxfordjournals.pan.a004868},
  journal = {\textit{Political Analysis}},
  shortjournal = {Polit Anal},
  author = {King, Gary and Zeng, Langche},
  urldate = {2025-09-21},
  date = {2001-06},
  langid = {english},
}

@inproceedings{sener_active_2018,
  title = {Active Learning for Convolutional Neural Networks: A Core-Set Approach},
  pages = {1--12},
  url = {https://openreview.net/forum?id=H1aIuk-RW},
  booktitle = {International Conference on Learning Representations (ICLR)},
  author = {Sener, Ozan and Savarese, Silvio},
  urldate = {2025-09-21},
  date = {2018-04},
  langid = {english},
}

@article{freund_selective_1997,
  title = {Selective Sampling Using the Query by Committee Algorithm},
  volume = {28},
  number = {2-3},
  pages = {133--168},
  issn = {0885-6125},
  url = {https://link.springer.com/article/10.1023/A:1007330508534},
  doi = {10.1023/A:1007330508534},
  journal = {\textit{Machine Learning}},
  shortjournal = {Machine Learning},
  author = {Freund, Yoav and Seung, H. Sebastian and Shamir, Eli and Tishby, Naftali},
  urldate = {2025-09-21},
  date = {1997-09},
  langid = {english},
}

\section*{Figure Legends}
\textbf{Figure 1:} RELEAP workflow. Patient-level inputs include clinical covariates for constructing proxy phenotypes, additional structured predictors for risk models, and a limited pool of costly reference labels. The RL-AL process adaptively selects patients to correct noisy proxy phenotypes into refined computable phenotypes, guided by a policy trained with PPO. The state vector summarizes short-term downstream performance trends, labeling budget, and strategy-level statistics. The policy outputs weights across active learning strategies (uncertainty, diversity, QBC), which drive the action of selecting patients for relabeling. Corrected phenotypes are then combined with additional structured predictors to train downstream risk prediction models. Model performance on validation data (e.g., AUC or C-index) is both logged as final results and fed back as the reward signal, closing the loop between downstream prediction and the RL-AL process.
\\
\textbf{Figure 2:} Coverage of smoking-related ICD codes (presence of $\geq 1$ smoking-related ICD per patient).
\\
\textbf{Figure 3:} Validation metrics vs.\ AL iteration in the logistic mode (mean $\pm$ 95\% CI over ten replications).
\\
\textbf{Figure 4:} C-index vs.\ AL iteration in the survival (Cox) mode (mean $\pm$ 95\% CI over ten replications).
\\
\textbf{Figure 5:} Validation metrics vs.\ AL iteration in the logistic mode (mean $\pm$ 95\% CI over ten replications) within male and female subgroups.

\end{document}


\section*{S1. Simulation Study}

\subsection*{Simulation Setting}
We evaluated the finite-sample performance of the proposed reinforcement learning–active learning (RL–AL) framework using synthetic data generated under the minimal causal structure $X_1 \to S_{\text{true}} \to S^{\ast}$ and $(S_{\text{true}}, X_2)\to Y$. The simulation setting is as follows.

\begin{itemize}
\item \textbf{Sample size $n$.} For each replicate, we generate a cohort of $n=1000$, split into 80\% training and 20\% validation sets stratified by $Y$, under a fixed labeling budget and batch size for active learning.
\item \textbf{Predictors $X_1$ and $X_2$.} Draw $X_1 \in \mathbb{R}^{d_{X1}}$ and $X_2 \in \mathbb{R}^{d_{X2}}$ independently from standard Gaussian distributions, providing phenotype-related and outcome-related signals, respectively.
\item \textbf{Automatic reference phenotype $S_{\text{true}}$.} Generate $S_{\text{true}}$ via a logistic link from $X_1$, with Gaussian perturbation applied to the linear predictor before passing through a sigmoid.
\item \textbf{Proxy phenotype $S^{\ast}$ (continuous).} Construct $S^{\ast}$ by adding Gaussian noise to $S_{\text{true}}$ and applying a sigmoid, mimicking a noisy ICD-based proxy label.
\item \textbf{Outcome $Y$.} Generate the binary outcome from a logistic model with a fixed coefficient for $S_{\text{true}}$ and random coefficients on $X_2$.
\end{itemize}

We compared random sampling, uncertainty, diversity, query-by-committee (QBC), and the proposed RELEAP under identical labeling budgets.

\subsection*{Simulation Results}
\begin{figure}[htbp]
  \centering
  \includegraphics[width=\textwidth]{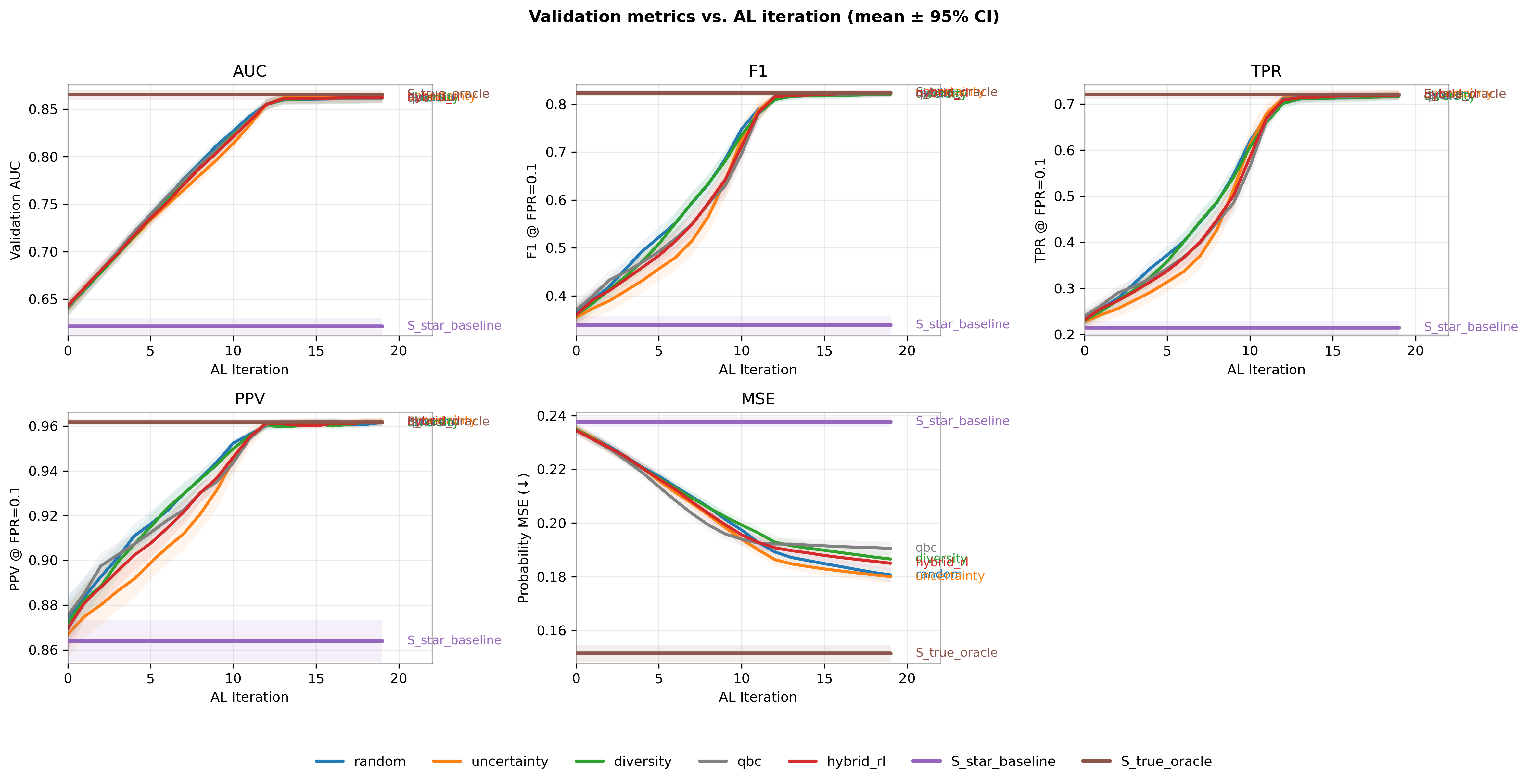}
  \caption{Simulation study: validation metrics across active-learning iterations (mean $\pm$ 95\% CI, $n=100$ runs).}
  \label{fig:sim-metrics}
\end{figure}

Figure~\ref{fig:sim-metrics} summarizes validation metrics across active-learning iterations ($n=100$ runs). All active-learning strategies improved over the noisy $S^{\ast}$ baseline and converged toward the oracle $S_{\text{true}}$ reference.

\FloatBarrier
\section*{S2. Active Learning Strategies}

We evaluated several active learning (AL) strategies used in RELEAP for selecting the most informative unlabeled patients to query during each iteration $t$.  
Let $\mathcal{L}_t$ denote the labeled set, $\mathcal{U}_t$ the unlabeled pool, and $Y$ the clinical outcome (e.g., lung cancer occurrence).  
Each sample $i \in \mathcal{U}_t$ is represented by a feature vector composed of its phenotype proxy $S$ and structured covariates $X_2$.

\subsection*{Uncertainty}
At iteration $t$, a logistic model $f_t$ is trained on $\mathcal{L}_t$ to predict the probability $\hat{p}_{t,i} = f_t([S_i, X_{2,i}]) = P(Y_i=1\mid S_i,X_{2,i})$ for each unlabeled patient $i$.  
The uncertainty score $H(\hat p_{t,i})$ quantifies the model’s confidence and is defined as the Shannon entropy:
\[
H(\hat p_{t,i}) = -\hat p_{t,i}\log \hat p_{t,i} - (1-\hat p_{t,i})\log(1-\hat p_{t,i}),
\]
where higher values indicate greater uncertainty (i.e., probabilities closer to 0.5).  
Patients with the highest uncertainty are prioritized for labeling.  
Here, $S_i$ denotes the current phenotype estimate, $X_{2,i}$ the structured covariate vector, and $\hat p_{t,i}$ the predicted event probability for patient $i$.

\subsection*{Diversity}
To ensure the labeled set covers diverse regions of the feature space, we compute pairwise cosine distances between each unlabeled patient $i$ and its $k=10$ nearest labeled neighbors in the standardized $[S, X_2]$ space.  
Let $d_{ij}$ denote the cosine distance between patient $i$ and labeled neighbor $j$.  
The diversity score for patient $i$ is defined as:
\[
D_i = \mathrm{mean}(d_{ij}) + \lambda \times \mathrm{std}(d_{ij}), \quad \lambda = 0.5,
\]
where the first term encourages global coverage and the second term favors patients in underrepresented regions.  
Patients with the highest $D_i$ values are selected.

\subsection*{Query-by-Committee (QBC)}
To capture model disagreement, a committee of $M=7$ logistic models $\{f_t^{(m)}\}_{m=1}^M$ is trained on bootstrap resamples of $\mathcal{L}_t$.  
Each model applies mild perturbations, including random variation in the $L_2$ regularization term and feature dropout with probability $p=0.1$.  
For each patient $i$, the QBC score quantifies the disagreement among models:
\[
Q_i = \mathrm{Var}_m\big[\hat p_{t,i}^{(m)}\big],
\]
where $\hat p_{t,i}^{(m)}$ is the predicted probability of $Y_i=1$ from model $m$.  
A small entropy stabilizer (weight 0.1) is added to prevent degenerate variance when predictions are extreme.

\subsection*{Random Baseline}
As a benchmark, we include a random sampling strategy that uniformly selects patients from $\mathcal{U}_t$ without using model-derived scores.  
This provides a lower-bound reference for evaluating the efficiency of informed querying strategies.

\FloatBarrier
\section*{S3. Reinforcement Learning Formulas}

The reinforcement learning (RL) component dynamically adjusts the weighting of these strategies.  
At iteration $t$, the system observes a state vector $s_t$, selects an action $\mathbf{w}_t$, and receives a reward $r_t$ based on the downstream model performance.

\subsection*{State}
The state $s_t$ aggregates information summarizing both model performance and data composition:
\[
s_t = [m_t, \mathbf{z}_t^{\text{strat}}, \mu_t^{(S)}, \sigma_t^{(S)}, \text{slope}(m), \text{var}(m), b_t],
\]
where  
- $m_t$ is the current downstream evaluation metric (AUC for classification or C-index for survival);  
- $\mathbf{z}_t^{\text{strat}}$ summarizes the per-strategy score distributions on $\mathcal{U}_t$ (median and 80th percentile for each strategy);  
- $\mu_t^{(S)}$ and $\sigma_t^{(S)}$ are the mean and standard deviation of phenotype values among labeled samples;  
- $\text{slope}(m)$ and $\text{var}(m)$ denote the short-term trend and variability of recent metric trajectories;  
- $b_t$ is the fraction of labeling budget remaining at iteration $t$.

\subsection*{Action}
The policy network outputs a normalized weight vector $\mathbf{w}_t$ assigning importance to each active learning strategy:
\[
\mathbf{w}_t = (w_t^{\text{unc}}, w_t^{\text{div}}, w_t^{\text{qbc}}), \quad w_i \ge 0, \ \sum_i w_i = 1.
\]
This vector linearly combines normalized scores from the uncertainty, diversity, and QBC strategies to determine which patients to query from $\mathcal{U}_t$.

\subsection*{Reward}
The reward $r_t$ measures the improvement in downstream model performance.  
Let $m_t$ denote the active metric (AUC or C-index) and $\bar m_t$ its moving average over the most recent $h$ iterations.  
The relative performance gain is defined as:
\[
g_t = \frac{m_t - \bar m_t}{1 - \bar m_t + \varepsilon},
\]
where $\varepsilon$ is a small constant for numerical stability.  
This gain is then scaled by two shaping factors:  
(1) a progress term $(1 + 2\,\text{prog}_t)$, where $\text{prog}_t$ is the fraction of budget already used, and  
(2) a trajectory term $\tau_t \in \{1.0, 1.2\}$ that upweights improving trends.  
The shaped reward is:
\[
R_t^{\text{raw}} = g_t (1 + 2\,\text{prog}_t)\tau_t,
\]
which is normalized online:
\[
r_t = \frac{R_t^{\text{raw}} - \mu_t}{\sigma_t + \varepsilon},
\]
where $\mu_t$ and $\sigma_t$ are the running mean and standard deviation of rewards.

\subsection*{RL Algorithm}
We optimize the policy using Proximal Policy Optimization (PPO), which learns a stochastic mapping $\pi_\theta(a\mid s)$ from states to actions by maximizing the expected discounted return:
\[
\max_\theta \ \mathbb{E}\Big[\sum_{t=1}^T \gamma^{t-1} R_t\Big],
\]
where $\gamma \in (0,1)$ is the discount factor.  
Experience tuples $(s_t, a_t, R_t, s_{t+1})$ are collected during active learning and used to update both the policy (actor) and value (critic) networks periodically.

\FloatBarrier
\section*{S4. Mathematical Formulation}

Let $\mathcal{A}=\{\text{uncertainty}, \text{diversity}, \text{QBC}\}$ denote the candidate strategy set.  
At iteration $t$, the agent observes state  
\[
s_t = [\mathbf{z}_{\text{strat}}(t), p^{\text{lab}}_+(t), \text{slope}(m), \text{var}(m), b_t],
\]
where $p^{\text{lab}}_+(t)$ is the proportion of positive samples in the labeled set.  
The action $\mathbf{w}_t \in \Delta^2$ represents the simplex weight vector over strategies.  
The shaped cumulative reward is:
\[
R_t = \alpha(m_t - m_{t-1}) + (1 - \alpha) \sum_{k=1}^{K} \gamma^{k-1} (m_{t+k-1} - m_{t+k-2}),
\]
where $\alpha \in [0,1]$ controls the balance between immediate and long-term performance gains, and $\gamma$ is the temporal discount factor.

\FloatBarrier
\section*{S5. Automatic Reference Phenotype Variants}

We compared several alternative definitions of the automatic reference phenotype ($S_{\text{true}}$), which serves as the refined phenotype label used to correct the noisy proxy ($S^{\ast}$).  
Each variant reflects a different combination of structured self-reported smoking information, traditional natural language processing (NLP)–derived concept identifiers, and large language model (LLM)–extracted mentions from clinical notes:

\begin{itemize}
    \item \textbf{Self-report + NLP CUIs:} Combines structured self-reported smoking status with Concept Unique Identifiers(CUIs) extracted from clinical notes using traditional NLP pipelines such as rule-based or dictionary-based methods.  
    \item \textbf{Self-report + LLM mentions:} Combines structured self-reported smoking status with smoking-related mentions extracted from clinical notes using a large language model (LLM).  
    \item \textbf{NLP CUIs only:} Uses concept identifiers extracted by traditional NLP methods without incorporating any structured self-reported information.  
    \item \textbf{LLM mentions only:} Uses smoking-related mentions identified by a large language model without relying on structured self-reported data.
\end{itemize}

\begin{figure}[htbp]
  \centering
  \includegraphics[width=\textwidth]{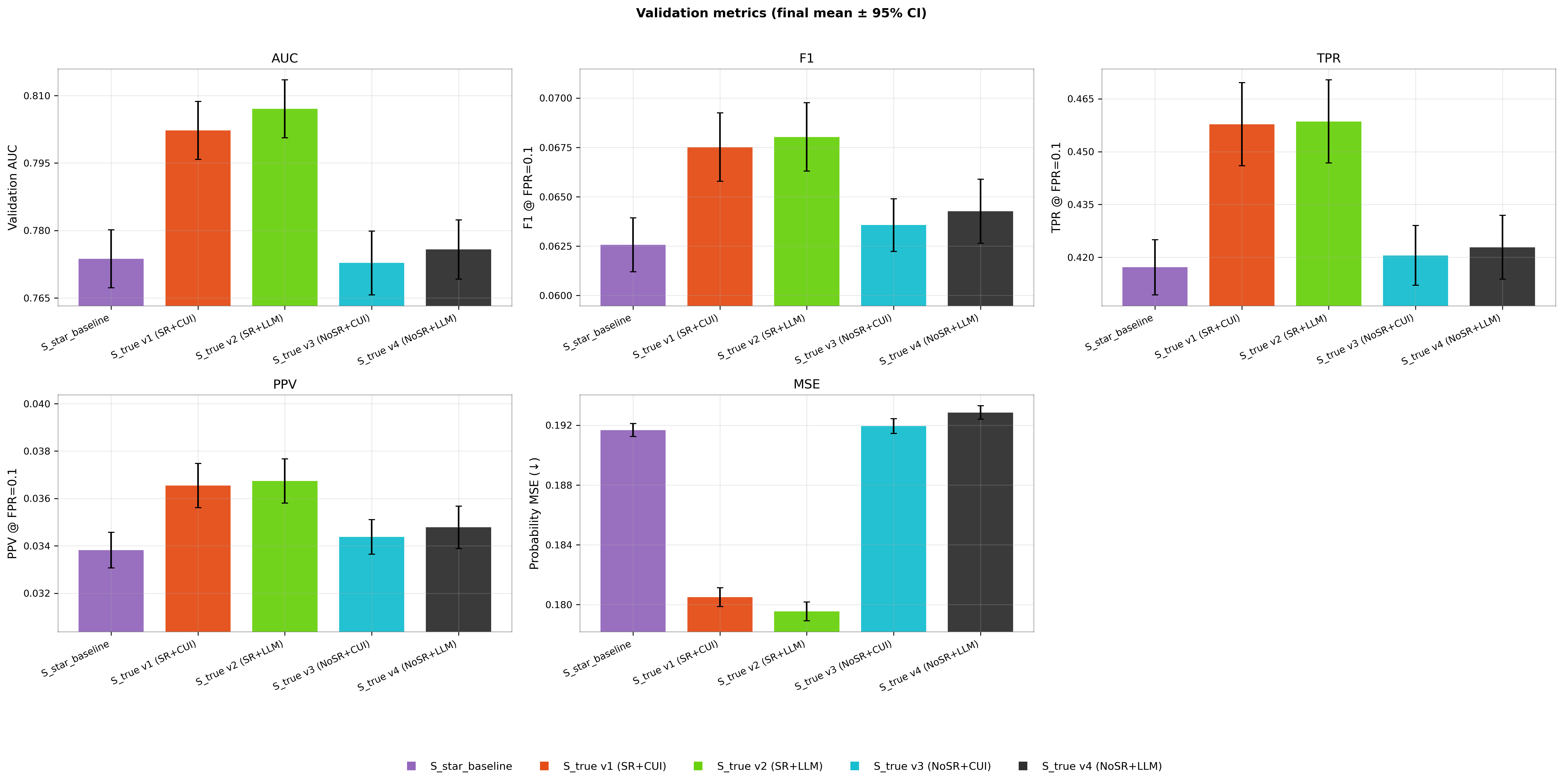}
  \caption{Comparison of Automatic Reference Phenotype variants ($S_{\text{true}}$ definitions) against the noisy proxy baseline ($S^{\ast}$). Bars show mean validation performance (AUC, F1 score, TPR, PPV, MSE) with 95\% CI error bars.}
  \label{fig:arp_variants}
\end{figure}

Figure~\ref{fig:arp_variants} summarizes validation performance across discrimination (area under the receiver operating characteristic curve, \textbf{AUC}), threshold-based metrics (F1 score, true positive rate [\textbf{TPR}], and positive predictive value [\textbf{PPV}]), and calibration (mean squared error of predicted probabilities, \textbf{MSE}).  

Across all evaluation metrics, the definitions that integrated structured self-reported information with either NLP- or LLM-derived text features achieved the highest predictive performance relative to the noisy proxy phenotype ($S^{\ast}$).  
By contrast, definitions relying solely on unstructured text—whether extracted by traditional NLP or LLMs—showed reduced accuracy and calibration.

\FloatBarrier